\documentclass[conference]{IEEEtran}
\IEEEoverridecommandlockouts
\usepackage{cite}
\usepackage{amsmath,amssymb,amsfonts}
\usepackage{algorithmic}
\usepackage{graphicx}
\usepackage{textcomp}
\usepackage{xcolor}
\usepackage{multirow}
\usepackage{pifont}

\newboolean{showcomments}
\setboolean{showcomments}{true}
\ifthenelse{\boolean{showcomments}}
 { \newcommand{\mynote}[2]{
      \fbox{\bfseries\sffamily\scriptsize#1}
        {\small$\blacktriangleright$\textsf{\emph{#2}}$\blacktriangleleft$}}}
        { \newcommand{\mynote}[2]{}}

\newcommand{\dcircle}[1]{\ding{\numexpr171 + #1}}

\def\BibTeX{{\rm B\kern-.05em{\sc i\kern-.025em b}\kern-.08em
    T\kern-.1667em\lower.7ex\hbox{E}\kern-.125emX}}
\begin{document}

\title{Letz Translate: Low-Resource Machine Translation for Luxembourgish\\

}

\author{\IEEEauthorblockN{Yewei Song\IEEEauthorrefmark{1},
Saad Ezzini\IEEEauthorrefmark{1}, Jacques Klein\IEEEauthorrefmark{1}, 
Tegawende Bissyande\IEEEauthorrefmark{1}, Clément Lefebvre\IEEEauthorrefmark{2}, and Anne Goujon\IEEEauthorrefmark{3} }
\IEEEauthorblockA{\IEEEauthorrefmark{1}\textit{Interdisciplinary Center for Security, Reliability and Trust (SnT)}, \textit{University of Luxembourg}\\
Email: \IEEEauthorrefmark{1}FirstName.LastName@uni.lu,\\
\IEEEauthorrefmark{2}\IEEEauthorrefmark{3}Banque BGL BNP Paribas, 50, avenue J.F Kennedy, L-2951 Luxembourg  \\
Email: \IEEEauthorrefmark{2}clement.c.lefebvre@bgl.lu,
\IEEEauthorrefmark{3}anne.goujon@bgl.lu}}
\maketitle

\author{\IEEEauthorblockN{Yewei Song}
\IEEEauthorblockA{yewei.song@uni.lu}
\and
\IEEEauthorblockN{Saad Ezzini}
\IEEEauthorblockA{saad.ezzini@uni.lu}
\and
\IEEEauthorblockN{Jacques Klein}
\IEEEauthorblockA{jacques.klein@uni.lu}
\and
\IEEEauthorblockN{Tegawendé F. Bissyandé}
\IEEEauthorblockA{tegawende.bissyande@uni.lu}
\and
\IEEEauthorblockN{Clément Lefebvre}
\IEEEauthorblockA{clement.c.lefebvre@bgl.lu}
\and
\IEEEauthorblockN{Anne Goujon}
\IEEEauthorblockA{anne.goujon@bgl.lu}
}

\pagestyle{plain}
\maketitle

\begin{abstract}
Natural language processing of Low-Resource Languages (LRL) is often challenged by the lack of data. Therefore, achieving accurate machine translation (MT) in a low-resource environment is a real problem that requires practical solutions. Research in multilingual models have shown that some LRLs can be handled with such models. 
However, their large size and computational needs make their use in constrained environments (e.g., mobile/IoT devices or limited/old servers) impractical.
In this paper, we address this problem by leveraging the power of large multilingual MT models using knowledge distillation. Knowledge distillation can transfer knowledge from a large and complex teacher model to a simpler and smaller student model without losing much in performance. We also make use of high-resource languages that are related or share the same linguistic root as the target LRL. For our evaluation, we consider Luxembourgish as the LRL that shares some roots and properties with German. We build multiple resource-efficient models based on German, knowledge distillation from the multilingual No Language Left Behind (NLLB) model, and pseudo-translation. We find that our efficient models are more than 30\% faster and perform only 4\% lower compared to the large state-of-the-art NLLB model.
\end{abstract}


\section{Introduction}

Natural language is a valuable heritage of mankind. It needs to be preserved, developed, and bridged with other languages. Natural language processing (NLP) has been shown to have a direct impact on the development and maintenance of natural languages, such as enabling the Neural Machine Translation of Hokkien, an unwritten language, to English~\cite{Hokkien}. Machine Translation (MT) is one of the most useful day-to-day applications of NLP that improves inter-cultural human interactions. 

Neural Machine Translation (NMT) has revolutionized the MT task in recent years. Recent advances in language models and transfer learning have allowed the MT task to scale up to a large extent by taking advantage of existing large parallel corpora. However, Low-Resource Languages (LRLs) lack such large corpora. Thus, making classic NMT models for LRLs is not possible. 
Although the recent Meta's No Language is Left Behind (NLLB)~\cite{Costa:22} multi-language translation model can handle 200 languages, its large size and computational needs make running the model in low resource environment (e.g., online services, mobile/IoT devices) impractical.

Motivated by addressing the discussed limitations, we propose resource-efficient translation models for LRLs using knowledge distillation and language-approaching techniques. Our proposed technique aims at achieving near state-of-the-art performance with minimal computational power. 
Knowledge distillation is concerned with using a large-scale multilingual teacher model (i.e., NLLB) to teach a student model the related task to achieve approximately the same performance with a low inference cost. The language-approaching technique adapts the resources of a high-resource language (HRL) that is related to the target LRL by performing token and expression replacement. Our approach aims to provide accurate translation models with low computational cost, few parallel data, and fast inference time. In our case study, our objective is to efficiently translate Luxembourgish (LB) to English (EN). So we consider LB as the LRL that is related to German (DE), HRL in our case. In our experiments, we found that our data-efficient distillation-based approach achieves better results compared to the use of large models either with larger parallel datasets for training or when augmenting the existing LB data using pseudo-translation from DE.







\section{Related Work}

In this section, we position our work to the existing literature on low-resource NMT that uses models or data from another HRL.

In their article~\cite{Ko:21}, Ko et al. proposed an adaptation method that combines large parallel HRL and monolingual LRL data to create an unsupervised translation model that can generalize across language groups. The authors found that their approach can achieve up to 36\% bilingual evaluation understudy (BLEU) score (24\% improvement compared to using the Spanish model) in translating Portuguese (an LRL) from and to English. Their approach relies on back-translation and adversarial domain adaptation techniques prior to a final fine-tuning step. The paper does not discuss the computational efficiency of the created models. Similar to our article, there are existing works that utilize an HRL that is similar to the target LRL (e.g., share the same root) to improve the NMT of the LRL from or to English.  Neubig et al.~\cite{Neubig:18} among others~\cite{Johnson:17,Zoph:16} used the technique of mixing parallel data from both LRL and HRL to train and improve the NMT of the LRL of interest. This technique is based on either using the HRL for pre-training and the LRL data for fine-tuning, or by mixing both datasets after balancing and augmenting the minority language data to result in one training set. The former proved to have a better performance. 

In contrast to the existing literature, in our work, we produce resource- and data-efficient models using knowledge distillation from a large multilingual model. We also compare the knowledge distillation to using a related HRL model, fine-tuning it on an LRL parallel set, and on a synthetic set created using pseudo-translation by transforming a large HRL parallel dataset. 

\begin{figure*}[ht]
\begin{center}
\includegraphics[width=0.88\textwidth]{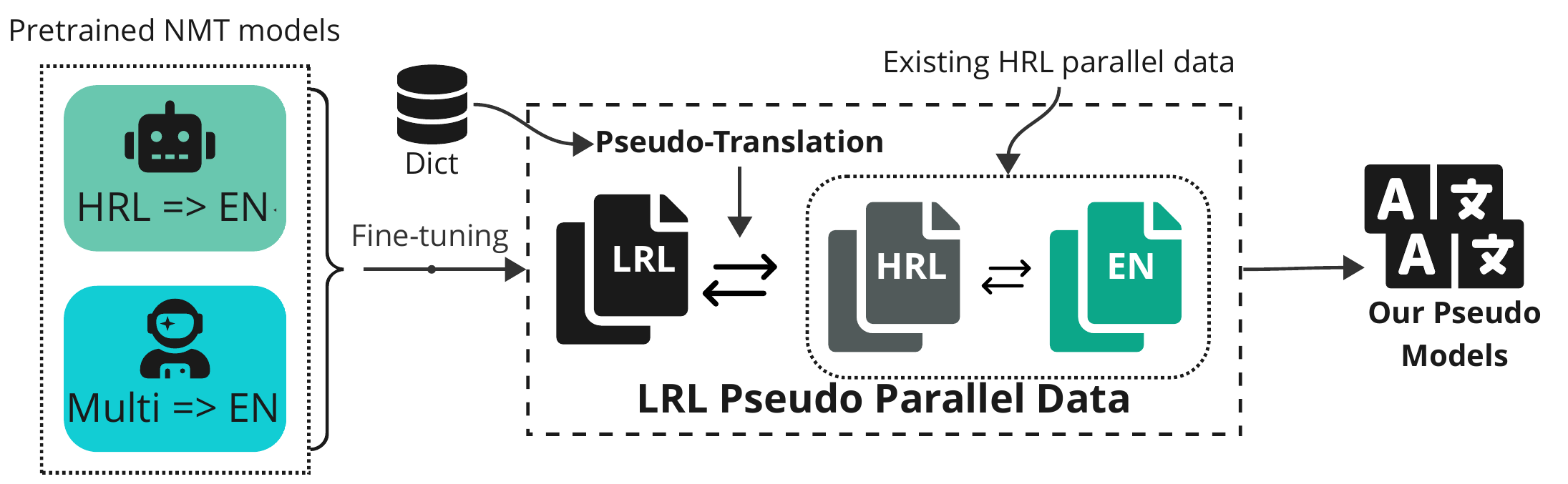}
\end{center}
\caption{The pipeline for obtaining our pseudo models} 
\label{fig:lrt} 
\end{figure*}

\begin{figure*}[ht]
\begin{center}
\includegraphics[width=0.88\textwidth]{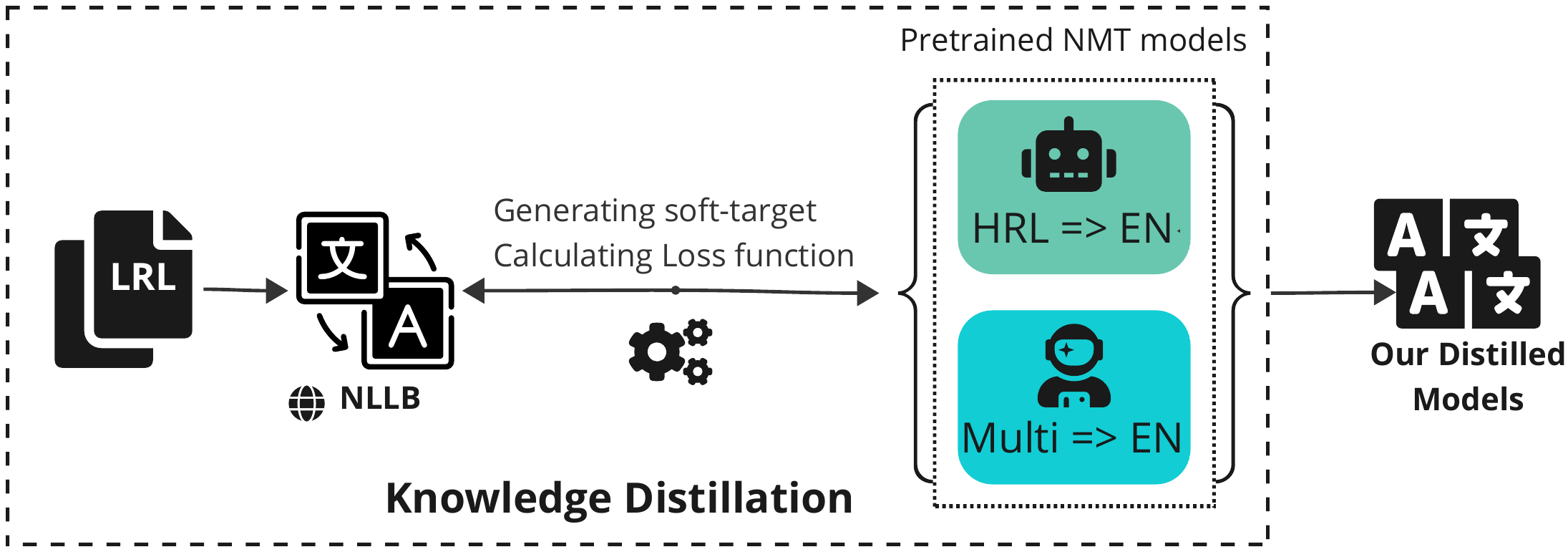}
\end{center}
\caption{The pipeline for obtaining our distilled models} 
\label{fig:lrt2} 
\end{figure*}

\section{Approach}

This section describes the main pipelines and various techniques that are used in this work.

\subsection{Problem Definition}

Our aim is to obtain a \textbf{fast} and \textbf{high-performing} translation model \dcircle{1} using a few data from an LRL and \dcircle{2} in an environment with limited computing power.

We will consider a translation model as \textbf{fast} if the model's inference time per instance is significantly smaller than the inference time required by state-of-the-art approaches such as NLLB.
As an example, the large NLLB model (with 3.3B parameters) has an inference time of 17 seconds per instance on a 48-core CPU server, which makes it relatively slow and impractical.
A translation model can be considered as \textbf{high-performing} if its translation scores are close to the state-of-the-art larger models.





As a first constraint, we will use limited LRL data to imitate the low resource case. For example, using a small \textbf{single-source corpus} of Luxembourgish (LB). As a second constraint, we will use our models in scenarios where computing resources are very restricted, such as mobile devices, intranets without GPU servers, etc. Our goal is to achieve a per-sentence inference time of less than 1 second on a CPU-only server.
 
\subsection{Two Ways for Getting Parallel Data}

In this paper, we compare the performance of the following two techniques: \dcircle{1} pseudo-translation and \dcircle{2} knowledge distillation, for LRL-to-EN translation models. We also conduct comparative experiments using our existing parallel corpus.

\textbf{Pseudo-Translation:} As in some previous work on low-resource language models such as LuxemBERT~\cite{lothritz2022luxembert}, and Translating Translationese~\cite{pourdamghani2019translating}, they use pseudo-translation to obtain acceptable quality sentences from another HRL that has a close linguistic or semantic relationship to the LRL. In our case, Luxembourgish is very similar to German in grammar and French (FR) in some vocabulary. 

\textbf{Knowledge Distillation:} In a NISP'2014 paper~\cite{hinton2015distilling}, Hinton et al. introduced the concept of Knowledge Distillation, which aims to migrate the knowledge learned from one large model or multiple models to another lightweight single model for easy deployment. In some cases, calculating the loss function using Soft-Target can produce a completely trained model without using any ground truth data. The concept of knowledge distillation is one of the original inspirations for this work. 

\subsection{Pipeline}

Figure~\ref{fig:lrt} outlines the pipeline for obtaining our \textbf{pseudo models}. In order to create our models from pseudo parallel data, we first need existing HRL-to-EN parallel data that can be used alongside a proper dictionary to perform the pseudo translation task and transform HRL sentences into pseudo LRL ones. Once the pseudo-parallel data is available, we use it to fine-tune existing pre-trained NMT models on the pseudo-LRL to the English set. Both HRL-to-EN and Multi-to-EN models can be used for this task.

In Figure~\ref{fig:lrt2}, we show the pipeline for creating our \textbf{distilled models}. For knowledge distillation, we first need an LRL corpus (not parallel) that can be fed to the large multilingual NLLB model to generate the soft target for HRL-to-EN or Multi-to-EN models with fewer parameters. The distillation process consists of passing the knowledge from a large complex teacher model to a smaller and simpler student model without losing much on the performance. Soft-target is used to minimize the error difference between both models.

Once our models are ready, an additional fine-tuning step can be performed on tiny but high-quality parallel data (LRL-to-EN). We will compare the two dimensions of using different first-round fine-tuning datasets and whether or not to use second-round fine-tuning. We also use the Ground-Truth fine-tuning set as a baseline.

Finally, we expect to obtain a Luxembourgish-to-English translation model with high computational speed, low computational resource consumption and performance close to that of a very large multilingual model. The best conclusion would be if our approach could be extended to all LRL-HRL translations. We will evaluate our created models in the next section.

\section{Evaluation}

\subsection{Experiment Design}
We rely on two datasets for our experiments. The first set is a pseudo-translated LB from DE-to-EN parallel data, obtained using the first pipeline in figure~\ref{fig:lrt}, and the second is the LB corpus to the generated soft target EN using the NLLB-Distilled model (600M). We needed to compare the translation performance on several different models. 

Many-to-one machine translation models have recently become increasingly popular, typically Romance to English, some of which include Luxembourgish and similarly other Frankish languages. We want to test the performance of our data on different pre-trained versions of similar machine learning models.
\subsection{Research Questions}

\textbf{RQ1. What is the most accurate solution for low-resource machine translation?} 

For RQ1, we want to assess the performance of the models that are based on pseudo-translation and knowledge distillation for our Luxembourgish use case, and compare them to our baselines.  

\textbf{RQ2. What is the best NMT model for LRL in terms of computational efficiency and inference time?}

For RQ2, we evaluate all models on the same hardware, to determine the best trade-off between model size, inference time, and translation accuracy.



\subsection{Implementation details and Availability}

The main libraries we used to create our models are Transformers, PyTorch, and Redis. We used the following pre-Trained Models: OPUS-MT-DE-EN, OPUS-MT-MUL-EN, NLLB-200-3.3B (NLLB-large) , and NLLB-200-DISTILLED-600M (NLLB-Distilled). We performed our training tasks on 1 Nvidia TESLA V100 32GB with Intel Xeon(R) CPU E5-2698, and run the inference on AMD EPYC 7552. The created model (Distilled training, 2-round fine-tuning, based on OPUS-MT-DE-EN) is available on the HuggingFace platform\footnote{https://huggingface.co/etamin/OPUS-TruX-Ltz-EN}.

\subsection{Data Collection}
In our evaluation, we use multiple datasets for different purposes, such as fine-tuning, pseudo-translation, and soft-target calculation. 

We consider the dataset provided by Gierschek in~\cite{gierschek2022detection} as our parallel dataset of \textbf{LB-DE-EN} sentences. She collected them from RTL Luxembourg\footnote{www.rtl.lu}, a Luxembourgish news website. We removed all sentence lengths shorter than 50 alphabets and longer than 500. It remains 110,720 parallel sentence triples. We call this dataset the \textbf{Ground-Truth}.

The \textbf{pseudo-translation} dataset is built by applying our collected bilingual dictionary on the ground-truth German to English set provided by Gierschek in~\cite{gierschek2022detection}. We collect a 51,617 words dictionary\footnote{github.com/Etamin/Ltz\_dictionary}, and apply Eifeler Regel (a special rule on words with suffix 'n' and 'nn') to German sentences\cite{Eifeler}.

The \textbf{soft target} for distilled fine-tuning, is generated by the NLLB-Distilled model\cite{nllb2022}. For this purpose, we use the Luxembourgish sentences from the ground truth (i.e., the dataset from Gierschek in~\cite{gierschek2022detection}), and no English sentences are used for mixing loss. It took us around 157 hours to generate the soft target. In our experiments, we used the smallest NLLB model in terms of size (NLLB-Distilled). Note that if we had used the largest version (NLLB-large), the experiments would require 350 additional hours processing time. 

We use the \textit{Flores-200}\footnote{https://huggingface.co/datasets/facebook/flores} train set for fine-tuning, and both the \textit{Flores-200} test set and the \textit{Tatoeba}\footnote{https://huggingface.co/datasets/tatoeba} dataset testing. 
Flores-200 is a widely used multilingual translation dataset between English and low-resource languages. It is an extended version of Flores101 released by Meta.
Flores-200 includes 997 pairs in the train set(for fine-tuning), 1017 pairs in the test set for test\cite{nllb2022}. Tatoeba is an open community-based collection of parallel sentences for 420 languages~\cite{10.1162/tacl_a_00288}. However, Tatoeba only has 306 Luxembourgish to English pairs.

\subsection{Pre-trained model and fine-tuning}
To build computationally efficient models, we have chosen two different pre-trained versions of OPUS-MT due to their cheap training cost, \textbf{OPUS-MT-DE-EN} and \textbf{OPUS-MT-MUL-EN}\cite{TiedemannThottingal:EAMT2020}.

For all fine-tuning tasks, we use the same training settings as follows. Max steps of 250,000, the learning rate is set to 2e-5 on AdamW, the weight decay is 0.01, and the batch size is 16. We also use the Cross-Entropy loss function.  

To verify the effect of pseudo-translation or knowledge distillation on the generalization ability of the models, we conduct a \textbf{second round} of fine-tuning in less than 500 steps using a subset of high-quality Flores-200. This process was designed to explore whether our data had the correct gradient for the model's ability to project from Luxembourgish to English. 

\subsection{Evaluation Metrics}

In our evaluation procedure, we use \textit{SacreBLEU} and \textit{ChrF++} to measure the performance of our models and the quality of the resulting translation compared to the ground truth.

\textbf{SacreBLEU} is an implementation variant of BLEU (Bilingual Evaluation Understudy) with the same objective of evaluating the quality of text translation by measuring the distance between the translated and ground-truth sentences. For simplicity, we will use the ``BLEU'' notation instead of SacreBLEU.

\textbf{ChrF++} is an improved version of ChrF which uses the F-score statistic for character n-gram matches. ChrF++ improves ChrF by adding n-grams of words to its computation.

\subsection{Experiments}

In this section, we outline the experiments we conduct to answer our research questions.

\subsubsection{Baseline models}
We consider three multilingual machine translation models that support Luxembourgish to English translation as baselines: NLLB-large (3.3B), NLLB-Distilled (600M), and M2M-100 (1.2M). 
In addition to these models, we consider two versions of OPUS-MT, one is multilingual as well (OPUS-MT-MUL-EN) that supports Luxembourgish, and the second is the German-to-English OPUS-MT-DE-EN model. For each version, we evaluate the models before and after fine-tuning the small Flores-200 Train Set. The selected baseline models are listed in Table~\ref{tab:baseline} which fine-tuned with Flores train set has an ``\textit{FT}'' suffix.

\begin{table*}
\caption{Baseline Model Performance}
\label{tab:baseline}
\begin{center}
\resizebox{.7\textwidth}{!}{%
\begin{tabular}{|l|ll|ll|l|}
\hline
Model (size) & \multicolumn{2}{l|}{\begin{tabular}[c]{@{}l@{}}Tatoeba\\ LB-EN\end{tabular}} & \multicolumn{2}{l|}{\begin{tabular}[c]{@{}l@{}}Flores 200\\ Test Set \end{tabular}}& Inference Time \\ \hline
Metrics (number of parameters) & \multicolumn{1}{l|}{BLEU} & ChrF++ & \multicolumn{1}{l|}{BLEU} & ChrF++ & second per sentence\\ \hline
NLLB-large (3.3B) & \multicolumn{1}{l|}{\textbf{56.019}} & \textbf{71.315} & \multicolumn{1}{l|}{\textbf{44.634}} & \textbf{68.981} & 16.99  \\ \hline
NLLB-Distilled (600M) & \multicolumn{1}{l|}{54.470} & 70.395 & \multicolumn{1}{l|}{38.589} & 64.424 & 3.26 \\ \hline
M2M-100 (1.2B) & \multicolumn{1}{l|}{20.679} & 35.054 & \multicolumn{1}{l|}{30.686} & 57.029 & 6.88 \\ \hline
OPUS-MT-MUL-EN (77M) & \multicolumn{1}{r|}{32.186} & 48.047 & \multicolumn{1}{l|}{19.819} & 46.89 & 0.51 \\ \hline
OPUS-MT-DE-EN (74M) & \multicolumn{1}{r|}{11.591} & 21.111 & \multicolumn{1}{l|}{5.370} & 24.591 & 0.28 \\ \hline
OPUS-MT-MUL-EN(FT) (77M) & \multicolumn{1}{r|}{36.763} & 55.044 & \multicolumn{1}{l|}{24.509} & 52.003 & 0.51 \\ \hline
OPUS-MT-DE-EN(FT) (74M) & \multicolumn{1}{r|}{23.941} & 39.215 & \multicolumn{1}{l|}{19.672} & 46.259 & 0.28 \\ \hline
\end{tabular}%
}
\end{center}
\end{table*}

In Table~\ref{tab:baseline}, we can see that NLLB-large takes around 17 seconds to translate one sentence. It has the largest number of parameters (3.3B) and therefore requires the most computational power. Obviously, this largest model has the best performance compared to other baselines. The M2M-100 \cite{fan2020beyond} is the second largest model with around 7 seconds per sentence execution time and a size of 1.2B parameters. However, it is only the fifth model in terms of accuracy, meaning that model size does not always correlate with performance. On the other hand, despite NLLB-Distilled being a lightweight version of NLLB and the second most accurate in our list, it still needs more than three seconds per sentence and has 600M parameters, which makes it a good baseline to compare against. OPUS-MT-based models can translate two to three sentences per second and have the smallest size (8 times smaller than NLLB-Distilled and 44 times than NLLB-large). These lightweight models are suitable candidates for our use case (online services that require fast inference time with low-computational resources).

\subsubsection{Our models}

\begin{table*}
\caption{Test results for all groups of translation}
\label{tab:results}
\begin{center}
\resizebox{0.7\textwidth}{!}{
\begin{tabular}{|lll|ll|ll|}
\hline
\multicolumn{3}{|l|}{Pre-Trained} & \multicolumn{2}{l|}{OPUS-MT-DE-EN} & \multicolumn{2}{l|}{OPUS-MT-MUL-EN} \\ \hline
\multicolumn{3}{|l|}{2nd Fine-tune} & \multicolumn{1}{l|}{Flores} & None & \multicolumn{1}{l|}{Flores} & None \\ \hline
\multicolumn{1}{|l|}{\multirow{2}{*}{Tatoeba}} & \multicolumn{1}{l|}{BLEU} & \multirow{4}{*}{Pseudo} & \multicolumn{1}{l|}{26.460} & 9.697 & \multicolumn{1}{l|}{37.553} & 14.480 \\ \cline{2-2} \cline{4-7} 
\multicolumn{1}{|l|}{} & \multicolumn{1}{l|}{ChrF++} &  & \multicolumn{1}{l|}{42.504} & 17.001 & \multicolumn{1}{l|}{55.697} & 26.401 \\ \cline{1-2} \cline{4-7} 
\multicolumn{1}{|l|}{\multirow{2}{*}{Flores}} & \multicolumn{1}{l|}{BLEU} &  & \multicolumn{1}{l|}{20.495} & 3.911 & \multicolumn{1}{l|}{22.351} & 5.301 \\ \cline{2-2} \cline{4-7} 
\multicolumn{1}{|l|}{} & \multicolumn{1}{l|}{ChrF++} &  & \multicolumn{1}{l|}{47.843} & 21.671 & \multicolumn{1}{l|}{49.823} & 26.478 \\ \hline
\multicolumn{1}{|l|}{\multirow{2}{*}{Tatoeba}} & \multicolumn{1}{l|}{BLEU} & \multirow{4}{*}{Distilled} & \multicolumn{1}{l|}{47.666} & \textbf{48.361} & \multicolumn{1}{l|}{46.051} & \textbf{47.906} \\ \cline{2-2} \cline{4-7} 
\multicolumn{1}{|l|}{} & \multicolumn{1}{l|}{ChrF++} &  & \multicolumn{1}{l|}{63.733} & \textbf{66.245} & \multicolumn{1}{l|}{\textbf{65.535}} & 65.508 \\ \cline{1-2} \cline{4-7} 
\multicolumn{1}{|l|}{\multirow{2}{*}{Flores}} & \multicolumn{1}{l|}{BLEU} &  & \multicolumn{1}{l|}{\textbf{29.790}} & 28.510 & \multicolumn{1}{l|}{\textbf{27.988}} & 25.521 \\ \cline{2-2} \cline{4-7} 
\multicolumn{1}{|l|}{} & \multicolumn{1}{l|}{ChrF++} &  & \multicolumn{1}{l|}{\textbf{57.618}} & 56.284 & \multicolumn{1}{l|}{\textbf{56.054}} & 53.897 \\ \hline
\multicolumn{1}{|l|}{\multirow{2}{*}{Tatoeba}} & \multicolumn{1}{l|}{BLEU} & \multirow{4}{*}{\begin{tabular}[c]{@{}l@{}}Ground\\ Truth\end{tabular}} & \multicolumn{1}{l|}{48.939} & 50.658 & \multicolumn{1}{l|}{47.813} & 48.444 \\ \cline{2-2} \cline{4-7} 
\multicolumn{1}{|l|}{} & \multicolumn{1}{l|}{ChrF++} &  & \multicolumn{1}{l|}{67.960} & 68.953 & \multicolumn{1}{l|}{67.057} & 67.824 \\ \cline{1-2} \cline{4-7} 
\multicolumn{1}{|l|}{\multirow{2}{*}{Flores}} & \multicolumn{1}{l|}{BLEU} &  & \multicolumn{1}{l|}{31.253} & 30.5207 & \multicolumn{1}{l|}{30.253} & 29.351 \\ \cline{2-2} \cline{4-7} 
\multicolumn{1}{|l|}{} & \multicolumn{1}{l|}{ChrF++} &  & \multicolumn{1}{l|}{58.899} & 58.032 & \multicolumn{1}{l|}{57.602} & 56.751 \\ \hline
\end{tabular}%
}
\end{center}
\end{table*}
As discussed earlier, for our evaluation, we consider two main approaches (i.e., Pseudo-translation and Knowledge distillation) applied to two OPUS-MT-based models, MUL-EN and DE-EN, as listed in Table~\ref{tab:results}. In this experiment, we compared the performance of our OPUS-MT-based models that are fine-tuned on the pseudo-translated set, the distilled set (obtained from distillation), and the whole ground truth. We also evaluate these models after a second fine-tuning on the Flores-200 train set.

As shown in Table~\ref{tab:results}, the pseudo-translation models perform better than M2M-100 and OPUS-MT-DE-EN models after the second fine-tuning, but have poor overall results as there was a large discrepancy between the pseudo-Luxembourgish sentences and the actual ground-truth, with BLUE scores lower than 35 points. These models are outperformed by OPUS-MT-MUL-EN and NLLB models on all datasets in Table~\ref{tab:baseline}.

The performance of the models obtained using distillation learning was very close to the models that use the 'ground-truth' parallel data for fine-tuning. On average, the difference is less than 2\% BLEU points. After the second round of fine-tuning with Flores-200, our model gained some improvement in some scenarios. 


In a cross-column comparison in Table \ref{tab:results}, the DE-EN model achieves better results after knowledge transfer to Luxembourgish, with an average advantage of around 1\% BLEU score. This may be due to the fact that some of the sentences in Luxembourgish are very similar to German, or it may be that some of the German knowledge has been retained in the fine-tuning. 

Based on the advantage of the DE-EN model over the MUL-EN model, we can guess that linguistically close bilingual models are more useful than multilingual models in transferring knowledge. However, to prove this, we need more experiments, such as testing the Limburgish (an LRL from the Netherlands) on Dutch (HRL) model and Romansh (an LRL from Switzerland) on the Italian (HRL) model.

Regarding time performance, as the inference time is not influenced by fine-tuning, this means that knowledge distillation brings the lightweight and fast inference benefits of OPUS-MT models with improved performance. According to Table \ref{tab:baseline}, our model inference speed (i.e., the inference time of OPUS-MT-MUL-EN) was 30 times faster than NLLB-large and 6 times faster than NLLB-Distilled.

\subsection{Answers to the RQs}

\textbf{RQ1.} What is the most accurate solution for low-resource machine translation?

The NLLB-large model is clearly ahead in terms of translation accuracy compared to other models or baselines. Among our solutions, the distilled models perform much better compared to the pseudo-models and the non-NLLB baselines (i.e., OPUS-MT and M2M in Table \ref{tab:baseline}). Moreover, they have a close performance to the same base models using large parallel data for fine-tuning. So, to answer this RQ, the best technique to provide accurate LRL translation using lightweight models is \textbf{knowledge distillation}.

\textbf{RQ2.} What is the best NMT model for LRL in terms of computational efficiency and inference time?

Given that the \textbf{OPUS-MT-DE-EN} model obtained using knowledge distillation, has the lowest inference time and model size, and has a slight performance edge over \textit{OPUS-MT-MUL-EN}, we can safely say that \textbf{OPUS-MT-DE-EN} is the best solution in this paper.  


%


\subsection{Discussion}

Firstly, our research demonstrates that using knowledge distillation can produce high-performance bilingual machine translation models using mega-multilingual models with only a single-side Luxembourgish corpus (RTL news or Wikipedia).
Second, we found that it is difficult to train effective unsupervised translation models even for using pseudo-translation from German to Luxembourgish (which are linguistically similar) when a lexicon and grammar adjustment is applied.

We also found that models of related languages have an advantage over multilingual models as a basis for transfer learning. To validate this observation, more comparative experiments on other languages are needed, e.g. Limburger, Lithuanian, Suomi, etc.

Additionally, after a manual check by a Luxembourgish native speaker, we realized that the Tatoeba dataset is too simple and lacks complexity. Flores-200, on the other hand, lacks noun diversity. The low quality of Tatoeba is expected as it is a community-based dataset.
A better evaluation dataset would have been welcomed. 

Finally, it should be noted that until a late stage of writing this article we were able to find out that the original OPUS-MT-MUL-EN model uses some Tatoeba data sets in its pre-pretraining, we are not sure the LB set was used. However, we do not observe an advantage of this model compared to OPUS-MT-DE-EN when evaluated on this dataset.


\section{Conclusion}
In this paper, we proposed two techniques to produce lightweight models for low-resource language (LRL) translation.
Our research demonstrates that high-performing low-resource language mini-models can be obtained using distillation learning based on large models. Our models are smaller, faster, and perform nearly as well as large multilingual NLLB models.

For future work, we plan to improve the pseudo-translation technique and test the knowledge distillation in other low-resource languages that have sparse parallel data. We also want to build and evaluate English-to-LRL translation models.


\bibliography{custom}
\bibliographystyle{IEEEtran}

\end{document}